\documentclass[3p, final]{elsarticle}

\usepackage{lineno,hyperref}
\usepackage{graphicx}
\usepackage{amsmath} 
\usepackage[FIGBOTCAP]{subfigure}
\usepackage{multirow}
\usepackage{xcolor}
\usepackage{ulem}
\usepackage{comment}

\journal{Journal}
\begin{document}

\begin{frontmatter}

\title{Unified Modeling of Unconventional Modular and Reconfigurable Manipulation System}

\author[mymainaddress,mysecondaryaddress]{Anubhav Dogra}
\date{}

\ead{2016mez0019@iitrpr.ac.in, anubhav.dogra@hotmail.com}

\author[mymainaddress,mysecondaryaddress1]{Sakshay Mahna}

\author[mymainaddress,mysecondaryaddress]{Srikant Sekhar Padhee}
\author[mymainaddress,mysecondaryaddress]{Ekta Singla}

\address[mymainaddress]{Indian Institute of Technology Ropar, Punjab, India, 140001}
\address[mysecondaryaddress]{Department of Mechanical Engineering}
\address[mysecondaryaddress1]{Department of Computer Science \& Engineering}

\begin{abstract}
Customization of manipulator configurations using modularity and reconfigurability aspects is receiving much attention. Modules presented so far in literature deals with the conventional and standard configurations. This paper presents the 3D printable, light-weight and unconventional modules: MOIRs' Mark-2, to develop any custom `\textit{n}'-Degrees-of-Freedom (DoF) serial manipulator even with the non-parallel and non-perpendicular jointed configuration. These unconventional designs of modular configurations seek an easy adaptable solution for both modular assembly and software interfaces for automatic modeling and control. A strategy of assembling the modules, automatic and unified modeling of the modular and reconfigurable manipulators with unconventional parameters is proposed in this paper using the proposed 4 modular units. A reconfigurable software architecture is presented for the automatic generation of kinematic and dynamic models and configuration files, through which, a designer can design, validate using visualization, plan and execute the motion of the developed configuration as required. The framework developed is based upon an open source platform called as Robot Operating System (ROS), which acts as a digital twin for the modular configurations. For the experimental demonstration, a 3D printed modular library is developed and an unconventional configuration is assembled, using the proposed modules followed by automatic modeling and control, for a single cell of the vertical farm setup.
\end{abstract}


\end{frontmatter}


\section{Introduction}
Current trend is moving towards the mass-customization of the products and the tools and has highlighted the need of custom design of the manipulators to be used in the collaborative environments like manufacting, maintenance, or services~\cite{althoff2019effortless, romiti2021toward}. Re-designing and re-manufacturing the custom robot from scratch is neither cost-effective nor time-efficient and that leads to the importance of modularity and reconfigurability aspects~\cite{chen2016modular,singh2016realization}. This not only aid in the customization of the robots but also reduces the downtime for the maintenance of the robotic manipulators, as the modules can be quickly replaced for repairing. Also, this incurs the challenges of automatic kinematic, dynamic and control modeling of the reconfigurable manipulators, for which computer-integrated solution is proposed in this paper. For non-repetitive tasks and for cluttered environments, where the conventional or standard configurations cannot withstand, a customized manipulator with non-parallel and non-perpendicular jointed configuration is required~\cite{singh2018modular}. Many researchers have contributed in this domain to achieve customization using modular and reconfigurable concepts. Reconfiguration has been presented using self and manually reconfigurable modular designs~\cite{liu2016survey,ahmadzadeh2016modular}, using lockable joints in~\cite{kereluk2017task,aghili2009reconfigurable} and using modular joints and links in~\cite{acaccia2008modular,song2011research,hong2017joint,liu2017modular,Mouliantis2019modulesreview}. Modular joints and links presented in these works are designed to get connected for standard configurations, such as planar chain, PUMA and SCARA configurations etc. A few researchers have also worked for unconventional configurations and modules, enabling to adapt the twist angles other than $0$ or $90^\circ$. Singh et al~\cite{singh2016realization} presented the design of modules with a twist unit to adjust the twist angles in two different ways. Modules had been presented in three variants based upon the size of the actuators used. Stravapodis et al.~\cite{moulianitis2016task} presented a pseudo joint module through which twist angle can be adjusted in discrete positions. Brandstotter et al.~\cite{brandstotter2015curved} presented rigid curved links to realize unconventional configurations. Recently, curvature adaptable links are presented in~\cite{brandstotter2018task} to get the desired configuration based upon the task. From standard to unconventional, modular designs have evolved according to the need of unconventional configurations to do complicated tasks or to be used in cluttered environments.\\

With these advancements, kinematic and dynamic modeling of the modular configurations is a challenge after re-configuration. Modeling of modular configurations are generally done by converting them into the known formulations such as DH parameters, Product of Exponential formula (POE),  screw theories, etc. Few recent works in this domain are presented such as, automatic kinematic modeling of the modular configuration using 2-DoF modules is performed using screw-theoretic formulations in~\cite{Modman2020}. Similar method is shown in~\cite{hossain2017enumeration, pan2013automatic,bi2007automated} by using graph theory, defining the Assembly Incidence Matrix (AIM) for the connections of the modules and calibrated using POE formula. An extension to standard DH parameters has been presented in~\cite{giusti2017fly,liu2020optimizing} for the automatic modeling of kinematics and dynamics. FEM based method for automatic modeling of modular robots is presented in ~\cite{bi2006automated}. DH parameters based modeling has been presented for standard modular configurations in~\cite{bi2007automated, djuric2006generalized}, where modular configurations are converted into the DH parameters for the forward and inverse kinematic computations. General orthogonal frames based kinematic modeling is presented in~\cite{mohamed2010module}. Brandstotter et al~\cite{brandstotter2015curved} uses the numerical algorithm developed by Husty and pfurner~\cite{husty2007new} for inverse kinematic solution of $6-DoF$ general configurations. Stravapodis et al~\cite{stravopodis2020rectilinear} uses product of exponential formula for forward and inverse kinematics of modular $3-DoF$ configuration. Majorly these works focus at the modeling of the standard configurations. Incorporating the parameters-adjustment within the modules are not discussed much. An approach for the kinematics of modular configuration is developed in this paper, by defining the three orthogonal frames. 
Followed by that, automatic reconstruction of desired configurations in a software environment is presented to visualize and validate the developed modular configurations. The virtual model is used not only in simulations but also to run the developed configuration in real-time - using the controllers - to accomplish the required tasks~\cite{vazquezvisual} and thus acts as digital twin. An efficient way to model the configuration for simulation and real-time control is through Unified Robot Description Format (URDF) file which can be incorporated in various robotic tool boxes such as Moveit, V-rep, Matlab, OpenRave etc. \cite{Kang2019}. Few works have been reported recently which incorporate the use of Robot Operating System (ROS)~\cite{ros} and URDF for the design of robotic configurations and can be seen in~\cite{ramos2018ontology,hernandez2017design,deng2017mobile,johannessen2019robot}. URDF can also be generated using some standard software such as SolidWorks, which has a plugin to convert the software assembly into robot description files. However, modeling the new configurations every time in computer aided design software is not time efficient. Thus, one of the challenges is to automatically generate the URDF for any \textit{n-DoF} modular configuration. 
Following observations can be outlined from the reported works.
\begin{enumerate}
	\item Only few reports are found which are dealing with the unconventional twist parameters, non-parallel and non-perpendicular jointed configurations, for modular manipulators.
	
	\item To develop the kinematic and the dynamic models of every different configuration needs thorough involvement, and a designer needs ample time analyzing any change while planning a configuration.
	
	\item Integration of automatic modeling and control of modular configurations along with the cluttered environment is essential during the optimal synthesis of the customized configurations, and that requires further exploration.
	
\end{enumerate}
Motivation of this work is to present a methodology for automatically generating the models of modular and reconfigurable manipulators even with non-parallel and non-perpendicular jointed configurations. These models can be visualized and aid in unifying the modeling and control of reconfigured manipulators through the proposed software architecture. The major contributions of this papers are, (a) Design of the modular library is presented which can be assembled in non-parallel and non-perpendicular fashion to develop the desired configuration with $n-$DoF. (b) Two methodologies to generate the automatic models (URDFs) of the given modular configurations with the proposed modular sequence and adaptive parameters are presented. First is using the direct modular assemblies and second is to convert from standard robot design parameters. (c) A reconfigurable software architecture is presented for automatic setting up of controllers and motion planner for the modular configurations in a given workspace. The methodology is integrated into an open source professional software for the simulation and real time control of the developed configurations.\\
The paper is organized as follows. Section~\ref{sec:modules} discusses the design of the modular architecture and respective features. Section~\ref{sec:modularconfigurations} defines the structural representation of the modules and corresponding kinematic computations. Automatic modeling of the configuration is presented in section~\ref{sec:automaticmodel}. The selected case studies of the modular configurations are presented in section~\ref{sec:results}, with the conclusion of the work in the last section.
\begin{figure}[t]
	\centering
	\subfigure[]{\includegraphics[width=1.25in]{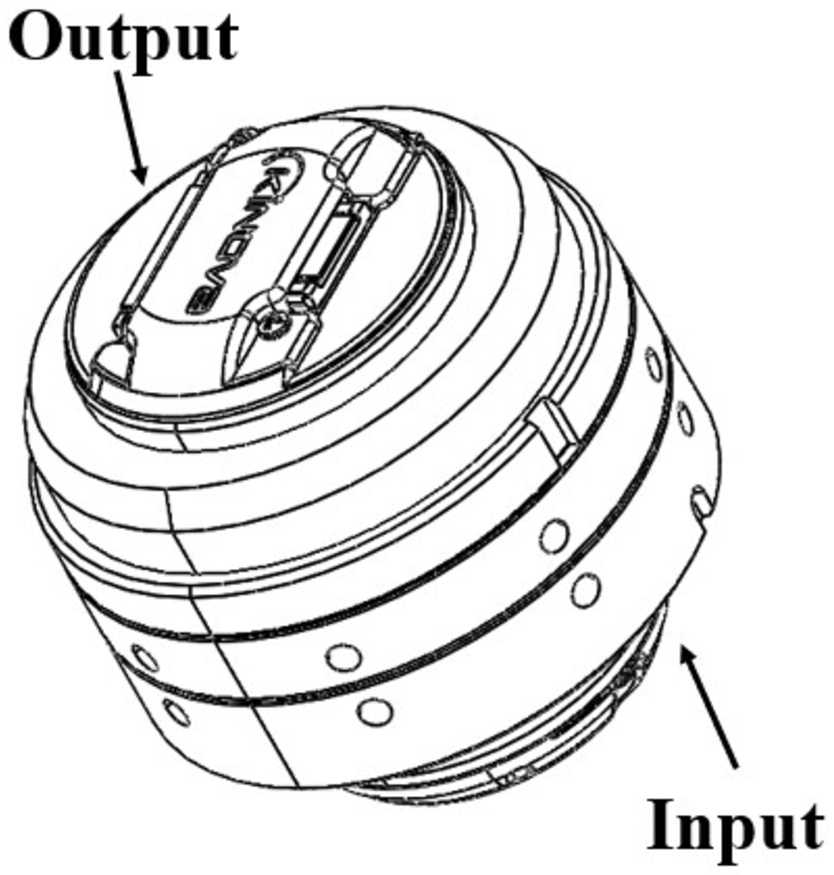}
	}
	~
	\subfigure[]{\includegraphics[width=2.5in]{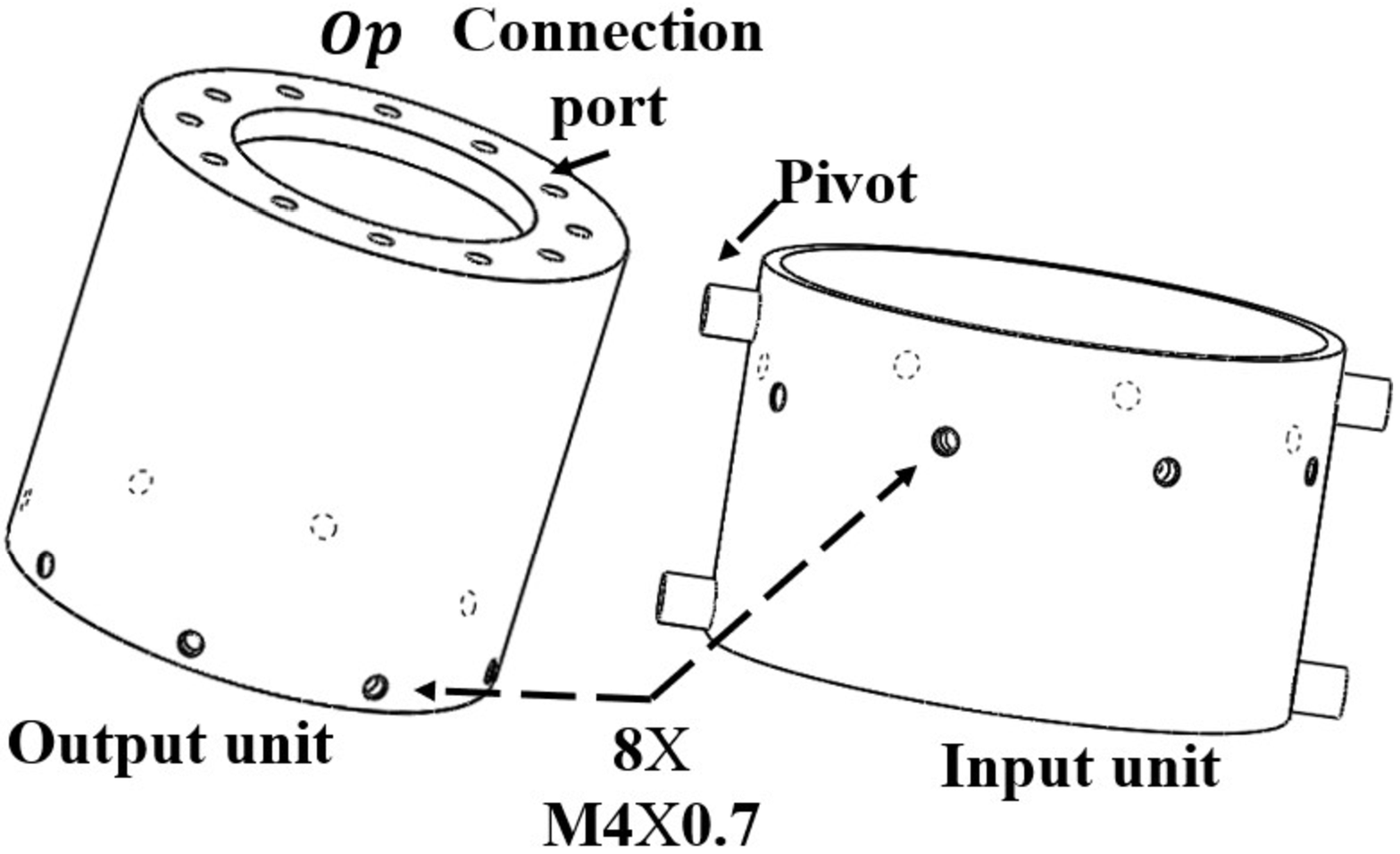}
	}
	~
	\subfigure[]{\includegraphics[width=2.7in]{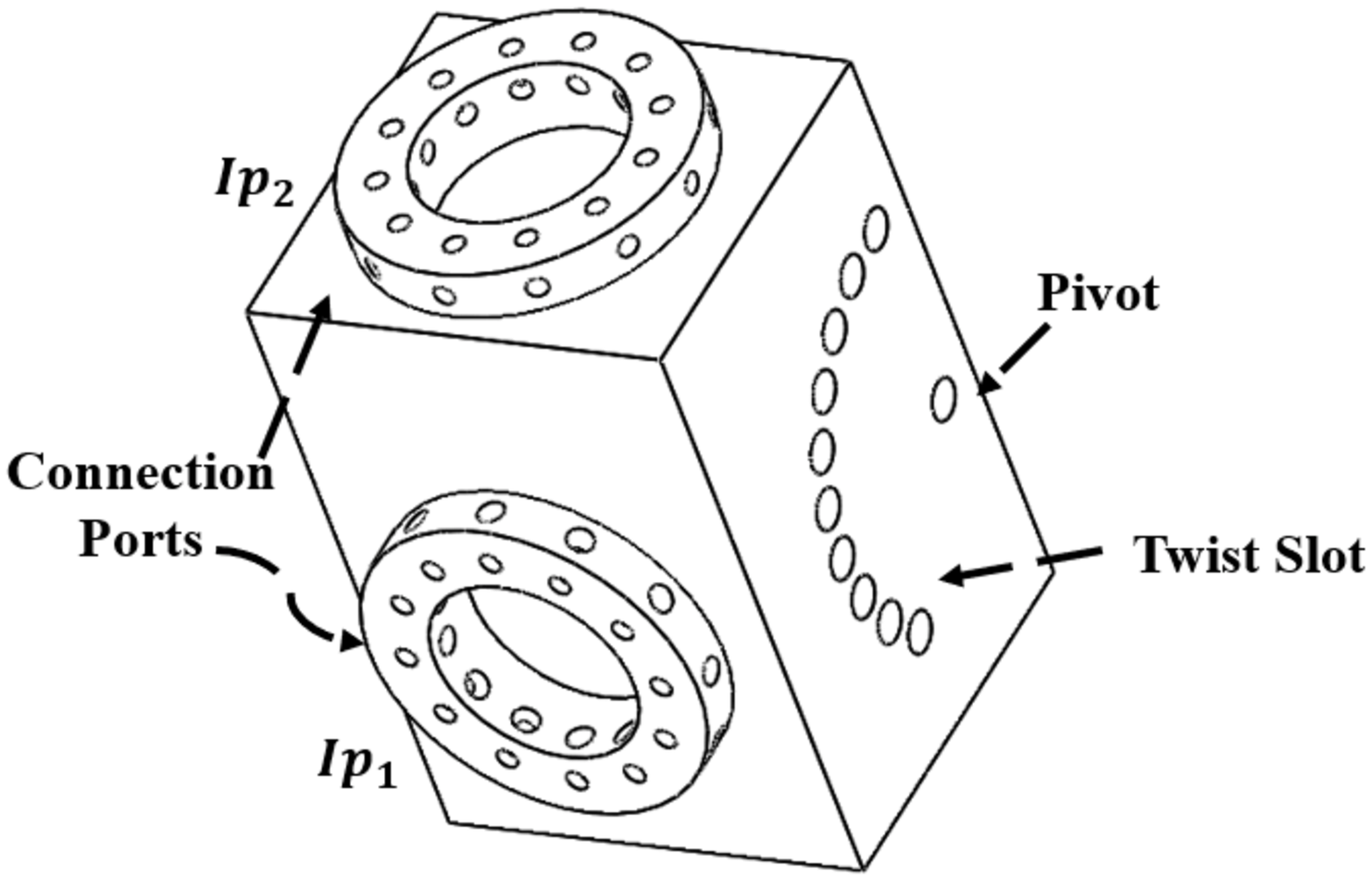}
	}
	~
	\subfigure[]{\includegraphics[width=1.5in]{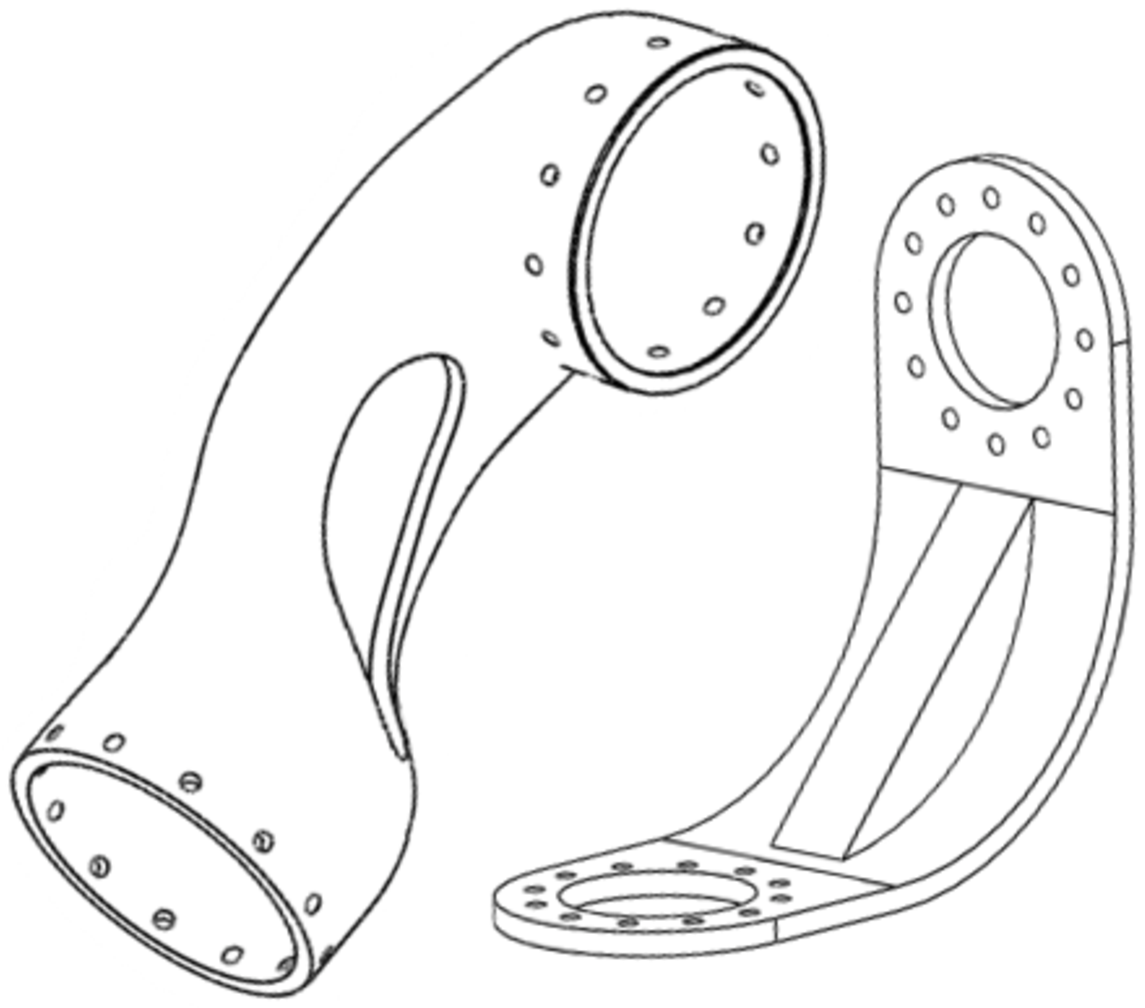}
	}
	\caption{Modular library designed to assemble customized modular and reconfigurable manipulators~\cite{dograJMD2021}. (a) Actuator, (b) Actuator casings, (c) Twist adjusting unit, (d) Link module.}
	\label{fig:joint_unit}
\end{figure}
\section{Adaptable modular library}\label{sec:modules}
In order to configure a customized manipulator, a modular library is presented containing joint modules and link modules, named as \textit{MOIRs' Mark-2}. The modules can be assembled for a $n-$DoF configuration and are also adaptable to the reconfigurable parameters. The modules are designed utilizing the optimal planning strategy~\cite{dograRobotica2021}, with respect to minimal dynamic torques, presented by the authors in~\cite{dograJMD2021}. The elements in the modular library are discussed as follows. 

\subsection{Joint modules}
A joint module consists of $3$ components as, unconventional twist unit, actuator and the actuator casings for input and output section of the actuators as shown in Fig~\ref{fig:joint_unit}. It is designed to incorporate the actuator and its adjustment in terms of connection with other modules and twist angle.
\subsubsection{Adaptive twist unit}
It is designed to incorporate the twist angles adjustment between the two frames according to DH convention. The twist adjustment can be done in two ways. One is the angle between the two adjacent intersecting joint axes, another is an angle between two adjacent skew joint axes. In this design, a pivotal extension from the input unit of casing is assembled with the pivot of twist adjusting unit, see Fig.~\ref{fig:joint_unit} (b). The twist unit has a discontinuous semi-circular slot with resolution of $15^\circ$ ranges from $0$ to $+90^\circ$ and $-45^\circ$ in both clockwise and anticlockwise direction as shown in Fig.~\ref{fig:joint_unit} (c). The input unit can rotate about pivotal axis in the twist unit to adjust the twist angle. Besides, there are two connection ports provided orthogonal to each other and are provided with $12$ holes with resolution of $30^\circ$ to adjust twist angle.
\subsubsection{Actuators}
To design the variants of the modules, \textit{KA-Series} actuators from \textit{Kinova}~\cite{kinova2019} are used, see Fig.~\ref{fig:joint_unit} (a). The KA-Series actuators consist of a motor and two disk-shaped sides as the input side and the output side. Under the action of the motor, the two sides rotate with respect to one another around their common central axis. The input side of the actuator along with the input linkage are fixed while the output side of the actuator rotates and thus moving the output linkage. The KA-series actuators are available in two variants as $KA-75+$ and $KA-58$. Specifications of these actuators are provided in Table~\ref{tab:actuator_specs}. Here, $\epsilon$ denotes the number of modules which can be carried by the same type of modules when assembled~\cite{dograJMD2021}. Each variant of the actuator connects to each other in a daisy chain using a single flat flex cable between two modules. The serial connection of the actuators with direct connectivity only to its adjacent modules, makes it perfect for modular application.

\subsubsection{Actuator casings}
Two casings are designed for the actuator for each of its side, see Fig.~\ref{fig:joint_unit} (b). The input casing is to be assembled with input section of actuator, which is relatively fixed. Thus, it has two pivotal extensions, which are assembled with the twist unit. The output casings is to be assembled with output section which rotates with respect to input section. On each of the casing, there are eight tapped holes which are used to fix these casings onto the actuator. Output casing has $12$ threaded holes. These $12$ holes gives the resolution of $30^\circ$ when assembling other modules to it. The twist slot in the module is used to adjust the desired twist angle between two adjacent joint axes.
\subsection{Link modules}
Link modules are passive modules to provide structural and the reachable aspects to a manipulator as shown in Fig.~\ref{fig:joint_unit} (d). The connection ports of the link are perpendicular to each other. It provides the length or offset parameters according to the type of connection.\\
The structural design aspects of the modules are not considered for the current study. Here, the focus is laid on the automatic modeling of the modular configurations.

\begin{table}[t]
	\caption{Technical specifications of actuators.}
	\begin{center}
		\label{tab:actuator_specs}
		\begin{tabular}{c c c}
			& & \\ 
			\hline
			\hline
			Quantity & KA-75+ (H) & KA-58 (L) \\
			\hline
			$W~(kg)$ & 0.57 & 0.357 \\
			$RPM$& 12.2 & 20.3 \\
			$\tau_{nom}~ (Nm)$& 12 &  3.6 \\
			$\tau_{max}~ (Nm)$& 30.5 &  6.8 \\	
			$\epsilon$&3&3\\	
			\hline
			\hline
		\end{tabular}
	\end{center}
\end{table}
\section{Topology representation for modular configurations}\label{sec:modularconfigurations} To identify each configuration, developed using the adaptable modules as discussed in section~\ref{sec:modules}, a topological representation of the modular composition is proposed. Joint modules are categorized into two variants based upon the actuator used in it. One with the $K-75+$ is named as Heavy (H) and other with $K-58$ is named as Light (L). Each joint module is having $3$ connection ports in total through which joint or link modules can be assembled. Two of them are input connection ports ($Ip_1,~Ip_2$) and the other is an output connection port ($Op$), as shown in Fig.~\ref{fig:joint_unit}. Joint module or the link module, say $(k-1)^{th}$ module is connected to the ${k^{th}}$ module at the input connections ports ($Ip$). And ${(k+1)^{th}}$ module is connected to the ${k^{th}}$ module through output connection port ($O_p$). Also, with two modules available as Heavy (H) and Light (L), set of rules for the physical assembly of any required configuration assembly are prepared as follows.
\begin{enumerate}
	\item Fix link 1: $H_{i}$ and link \textit{n}: $L_{n}$.
	\item Connect $H_{i}$~$\rightarrow$ $H_{i+1}$ at ($Ip_{1_{i+1}}~or~Ip_{2_{i+1}}$) or $H_{i}$ $\rightarrow$ $L_{i+1}$ at ($Ip_{1_{i+1}}~or~Ip_{2_{i+1}}$).
	\item $L_{i}$ $\rightarrow$ $L_{i+1}$ at ($Ip_{1_{i+1}}~or~Ip_{2_{i+1}}$).
	\item Refer $\epsilon$ for maximum number of each module from Table~\ref{tab:actuator_specs}.
	\item Link modules are to be used to fulfil respective requirement.
\end{enumerate}
Here $i=1:n$, with \textit{n} as number of DoF.
A joint module can be connected to the next joint module with or without the link module in between. $\epsilon$ is the parameters which defines how many number of modules of one kind can be carried by the same module itself. This study has been conducted based upon the worst torque analysis of the joint torques for each kind of module by the authors in~\cite{dograJMD2021}. An  \textit{n-DoF} manipulator will be the sequence of H and L modules with considerations of the above mentioned rules. Some of the exemplary configurations are shown in Sections~\ref{sec:types}-~\ref{sec:type3}.
\subsection{Frames assignment: modular kinematics}\label{sec:frames}
\begin{figure} 
	\centering
	\subfigure[]{\includegraphics[width=1.1in]{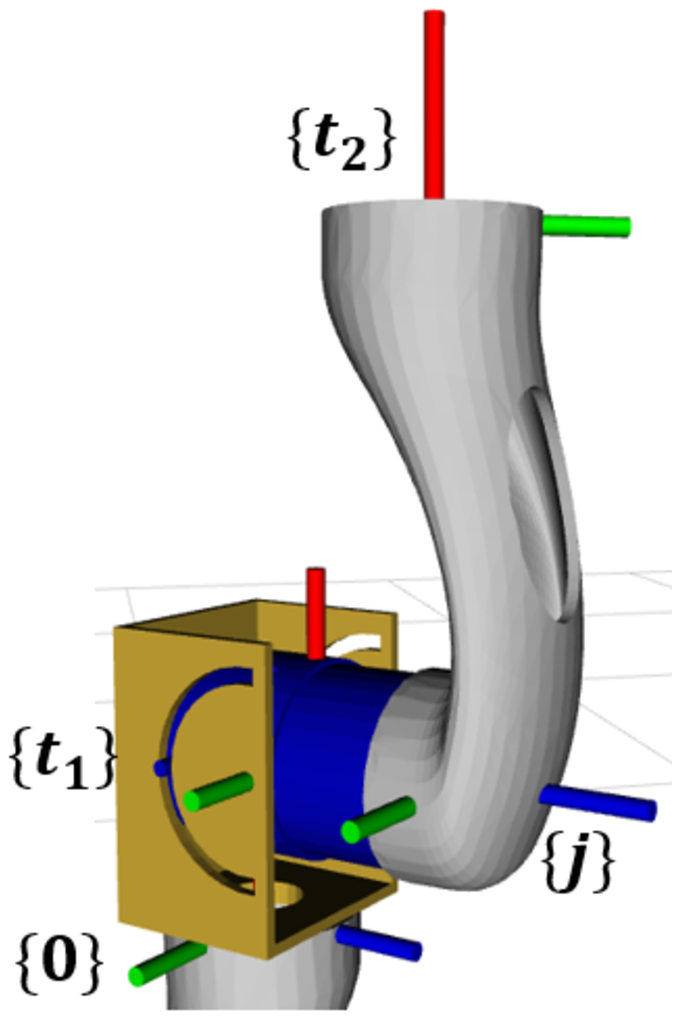}}
	~
	\subfigure[]{\includegraphics[width=1in]{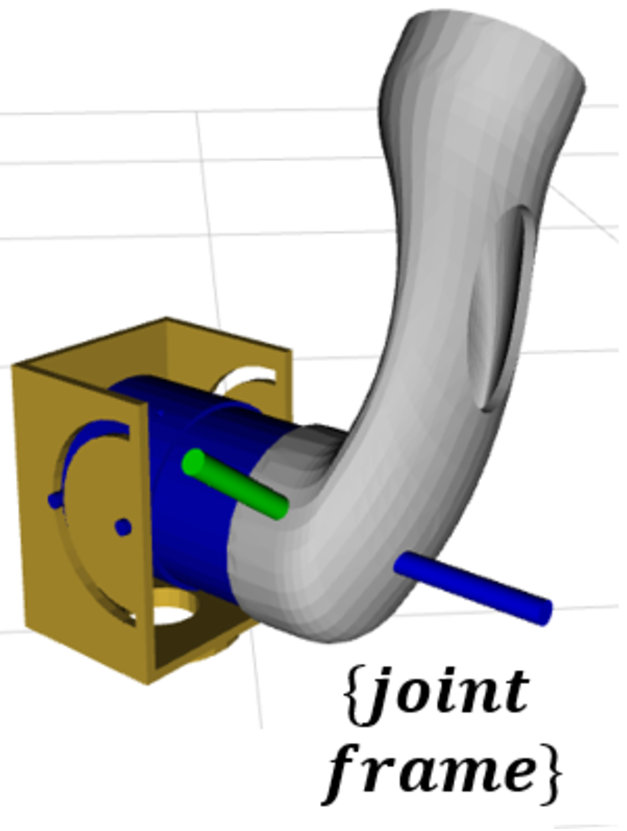}}
	~
	\subfigure[]{\includegraphics[width=1in]{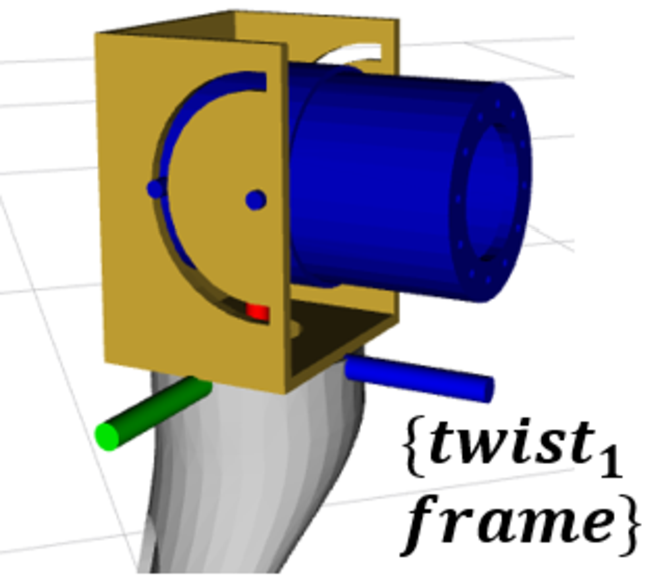}}
	~
	\subfigure[]{\includegraphics[width=1.1in]{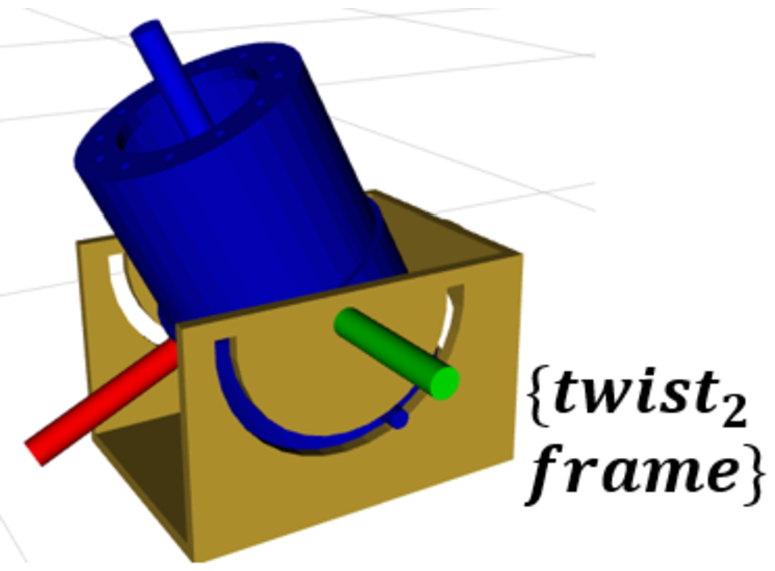}}
	\caption{(a) $3$ Frames associated to modular components, (b) Blue axis represents the z-axis of joint frame and is located on the actuator axis (b) Red axis represents the x-axis of twist$_1$ frame located at the connection port (c) Green axis represents the y-axis of the twist$_2$ frame located inside the twist adjustment module.}
	\label{fig:frames}
\end{figure}

\begin{figure} 
	\centering
	\subfigure[]{\includegraphics[width=0.8in]{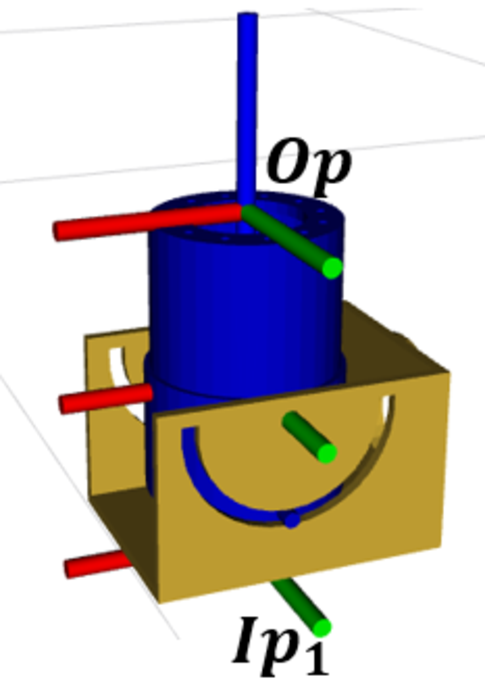}
	}
	~
	\subfigure[]{\includegraphics[width=1.1in]{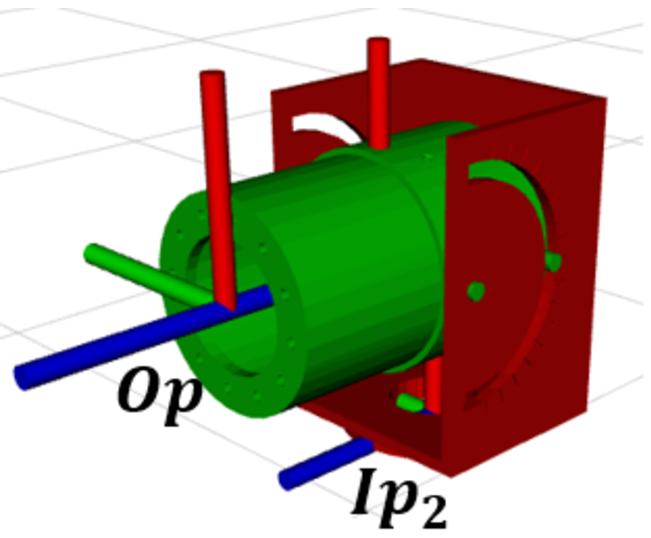}
	}
	~
	\subfigure[]{\includegraphics[width=1.3in]{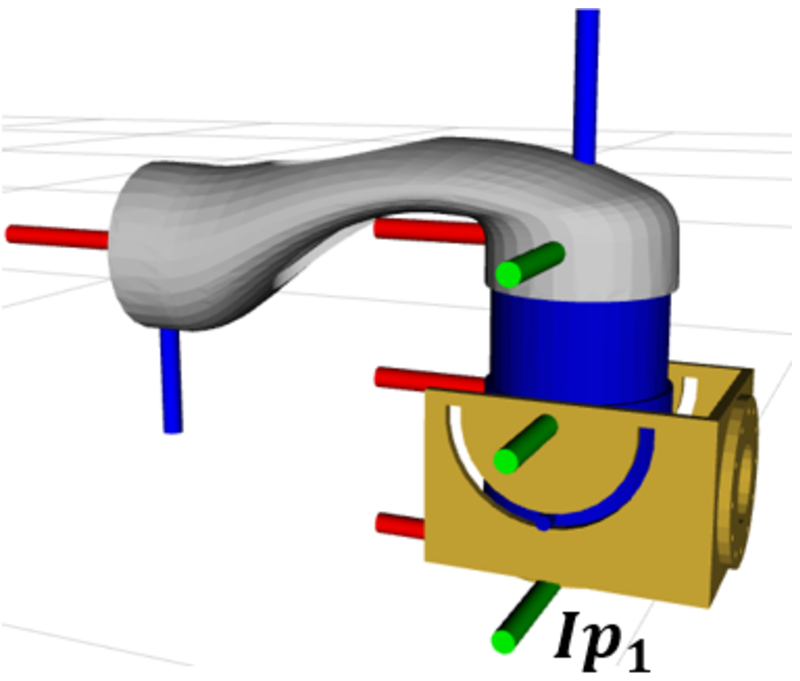}
	}
	~
	\subfigure[]{\includegraphics[width=0.8in]{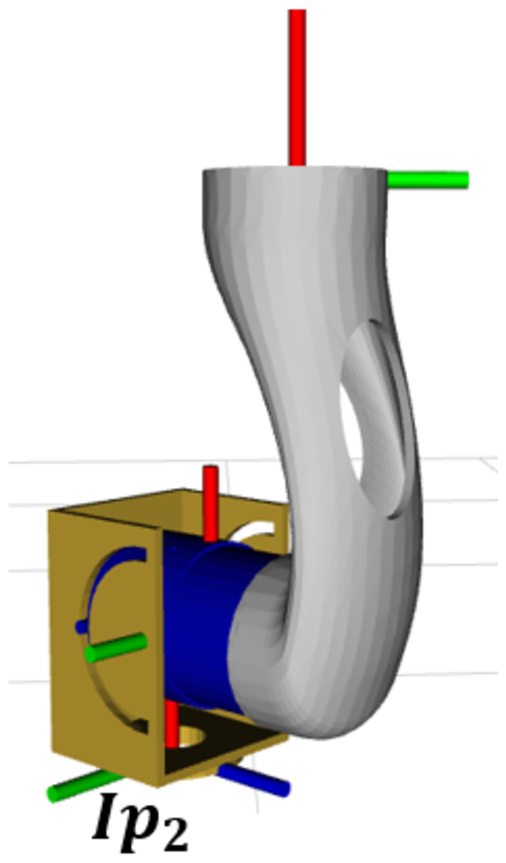}
	}
	\caption{The 4 types of modular units as (a) $H^1$ or $L^1$, (b) $H^2$ or $L^2$, (c) $H^3$ or $L^3$ and (d) $H^4$ or $L^4$.}
	\label{fig:4types}
\end{figure}

For the kinematic modeling of the modular configuration, and adjustment of the required parameters in the modules such as twist angle, joint angle, and link length, three orthonormal frames are associated to the modular links as shown in Fig.~\ref{fig:frames}. In this paper, red, green and blue axes represents $x$, $y$ and $z$ axes.
\begin{enumerate}
	\item Joint frame: Joint frame is a rotating frame, with variable as joint angle, about its own $z-$axis (blue) as shown in Fig.~\ref{fig:frames} (b). It is also an attaching frame for the connection of the modules with the $O_p$ port of the joint module. 
	
	\item Twist$_1$ frame: It connects previous modules with the $Ip_1$ port of the joint module. Adjustment of the type II twist, referred in section~\ref{sec:type2}, is done using this frame,  by rotating about the $x-$axis (red) of twist$_1$ frame, as shown in Fig.~\ref{fig:frames} (c). 
	
	\item Twist$_2$ frame: It lies inside the joint modules on the pivot axis with origin at the center of this axis. Type III twists, referred in section~\ref{sec:type3}, are adjusted using this frame by rotating the actuator casings about $y-$axis (green) of twist$_2$ frame as shown in Fig.~\ref{fig:frames} (d).
	
\end{enumerate}
The positions of these frames are fixed with respect to each other within the modules but the orientation variables (about red/green/blue) do get adjusted during reconfiguration.\\
In a serial chain with $n-DoF$ and $(n+1)-$joint frames, the $n$th frame can be represented in the base frame ($0$th) using the following equation,

\begin{equation}
	{}^{n}_{0}\textbf{A}={}^{1}_{0}\textbf{A}~~{}^{2}_{1}\textbf{A}~~.~.~.~~{}^{n}_{n-1}\textbf{A},
\end{equation}
which can be expanded for the three frames defined as
\begin{equation}
	\begin{split}
		{}^{n}_{0}\textbf{A}=\{{}^{t_1}_{0}\textbf{A}~~{}^{J}_{t_1}\textbf{A}~~{}^{t_2}_{J}\textbf{A}\}_{1}~~\{{}^{t_1}_{1}\textbf{A}~~{}^{J}_{t_1}\textbf{A}~~{}^{t_2}_{J}\textbf{A}\}_{2}.~.~.\\ .~.~.~\{{}^{t_1}_{n-1}\textbf{A}~~{}^{J}_{t_1}\textbf{A}~~{}^{t_2}_{J}\textbf{A}\}_{n}.
	\end{split}
\end{equation}
This can be simplified as
\begin{equation}
	{}^{n}_{0}\textbf{A}=\prod_{i=1}^{n}\{{}^{t_1}_{i-1}\textbf{A}~~{}^{J}_{t_1}\textbf{A}~~{}^{t_2}_{J}\textbf{A}\}_{i},
\end{equation}
where  
\begin{equation}
	{}^{t_1}_{i-1}\textbf{A}= \begin{bmatrix}\cos(\alpha_{t_2}) ~~& 0~~ & \sin(\alpha_{t_2})~~ & x_{01}\\
		0 ~~& 1~~ & 0~~ & 0\\
		-\sin(\alpha_{t_2})~~ & 0~~ & \cos(\alpha_{t_2})~~ & z_{01}\\
		0~~ & 0~~ & 0~~ & 1\end{bmatrix};
\end{equation}

\begin{equation}	 	
	{}^{J}_{t_1}\textbf{A}=\begin{bmatrix}\cos(\theta) ~~& -\sin(\theta)~~ & 0~~ & 0\\
		\sin(\theta) ~~& \cos(\theta)~~ & 0~~ & 0\\
		0~~ & 0~~ & 1~~ & z_{12}\\
		0~~ & 0~~ & 0~~ & 1\end{bmatrix};
\end{equation}

\begin{equation}
	{}^{t_2}_{J}\textbf{A}=\begin{bmatrix}1~~ & 0~~ & 0~~ & x_{23}\\
		0~~ & \cos(\alpha_{t_1}) ~~& -\sin(\alpha_{t_1})~~ & 0\\
		0~~ & \sin(\alpha_{t_1})~~ & \cos(\alpha_{t_1})~~ & 0\\
		0~~ & 0~~ & 0~~ & 1\end{bmatrix}.
\end{equation}
Here, $i$ is the number of joint modules in the modular assembly which is equal to the number of degrees-of-freedom (DoF). ${}^{t_1}_{i-1}\textbf{A}$ represents the transformation of the twist$_1$ frame with respect to the base frame or the frame at one of the connection port of the joint module. ${}^{J}_{t_1}A$ represents the transformation of the joint frame with respect to the twist$_1$ frame, and ${}^{t_2}_{J}A$ represents the transformation of the twist$_2$ frame with respect to the joint frame. $\alpha_{t_1}$ and $\alpha_{t_2}$ are the twist$_1$ and twist$_2$ parameters respectively. Length parameters in these transformation matrices, say $x_{01}$, $z_{01}$, $z_{12}$ and $x_{23}$, are the constant values as per geometrical design constraints. For the modular assembly, few of the transformation matrices will be the identity matrices if the corresponding module is not used.\\ 
It could be quite cumbersome to visualize the modular configuration merely using these frames. Hence, for the automatic generation of mathematical models of the developed configurations, four modular units are proposed based upon the connection ports on the joint modules, as shown in Fig.~\ref{fig:4types}. These are valid for both `H and `L' variant of the joint modules. These can be enumerated as, `$H^k$' associated to Heavy (H) modules and `$L^k$' associated to Light (L) modules, where $k~\in~1:4$. This includes all four possibilities associated to each type of module.
\begin{enumerate}
	\item `$H^1$' or `$L^1$' is used when a joint module is to be connected through $Ip_1$. The concatenated transformation matrix for this modular unit representing joint frame to the base frame (of the modular unit) is
	\begin{equation}
		\textbf{A}^1={}^{t_1}_{0}\textbf{A}~~{}^{J}_{t_1}\textbf{A}.
	\end{equation}
	Here the values of $x_{01}$, $z_{01}$, and $z_{12}$, are $0~m$, $0.074~m$, and $0.073~m$, respectively.
	
	\item `$H^2$' or `$L^2$' represents the corresponding joint module connected through $Ip_2$.
	In this case too, the concatenated transformation matrix representing joint frame to the base frame (of the modular unit) is
	\begin{equation}
		\textbf{A}^2={}^{t_1}_{0}\textbf{A}~~{}^{J}_{t_1}\textbf{A}.
	\end{equation}
	However, in this case, the values of $x_{01}$, $z_{01}$, and $z_{12}$, are $-0.0297~m$, $0.075~m$, and $0.073~m$, respectively. 
	
	\item `$H^3$' or `$L^3$' is used when a joint module is connected through $Ip_1$ and it is having a link module attached at $Op$. In this case, concatenated transformation matrix representing the twist$_2$ frame at the end of the link with respect to the base frame of modular unit is
	\begin{equation}
		\textbf{A}^3={}^{t_1}_{0}\textbf{A}~~{}^{J}_{t_1}\textbf{A}~~{}^{t_2}_{J}\textbf{A}.
	\end{equation}
	Here, the values of $x_{01}$, $z_{01}$, $z_{12}$, and $x_{23}$, are $0~m$, $0.074~m$, $0.073~m$, and $0.22~m$, respectively. 
	
	\item `$H^4$' or `$L^4$' represents a joint  module connected through $Ip_2$ and having a link module at $Op$. In this case concatenated transformation matrix representing the twist$_2$ frame with respect to the base frame of modular unit is
	\begin{equation}
		\textbf{A}^4={}^{t_1}_{0}\textbf{A}~~{}^{J}_{t_1}\textbf{A}~~{}^{t_2}_{J}\textbf{A}.
	\end{equation}
	In this case, the values of $x_{01}$, $z_{01}$, $z_{12}$, and $x_{23}$, are $-0.0297~m$, $0.075~m$, $0.073~m$, and $0.22~m$ respectively. 
\end{enumerate}
\subsection{Modular configuration types}
For any given task and the environment, any type of configuration could be needed, say, planar or standard spatial manipulator. Any required configuration can be formed using these modular units and can be categorized in 3 major types as follows. 

\begin{figure}[t]
\centering
	\subfigure[]{\includegraphics[width=2.2in]{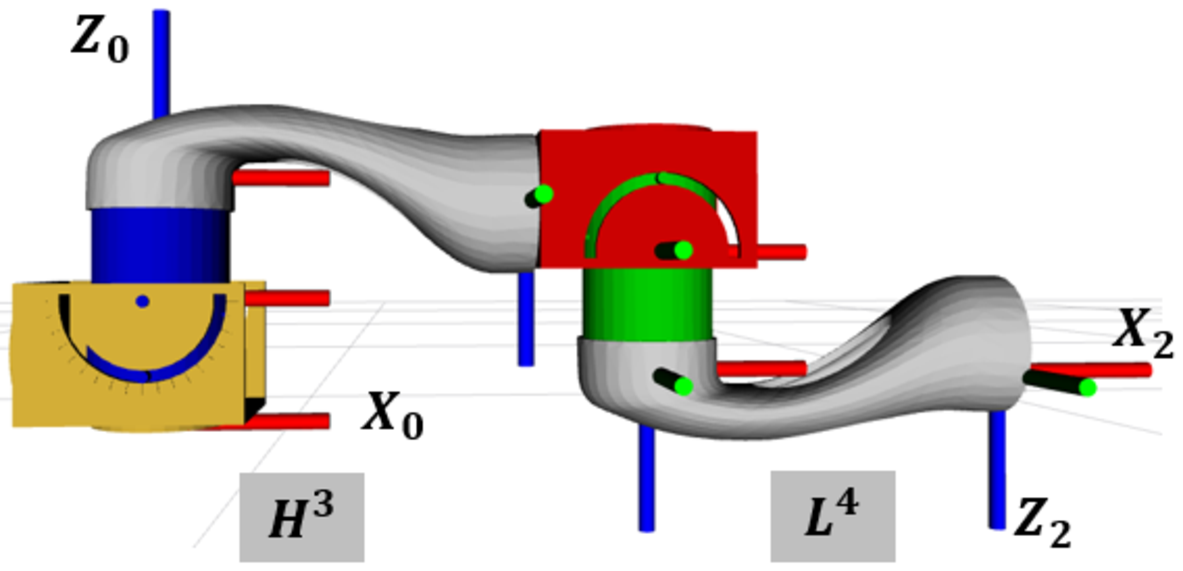}
	}
	~
	\subfigure[]{\includegraphics[width=1.25in]{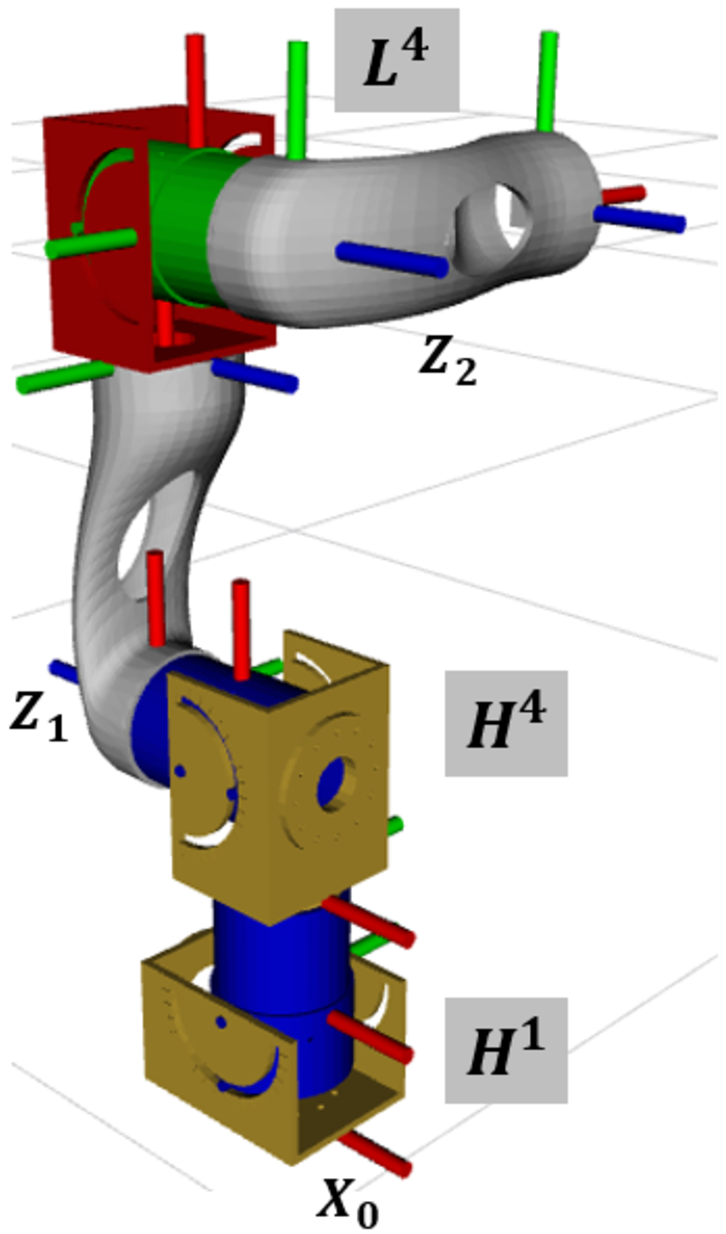}
	}
	~
	\subfigure[]{\includegraphics[width=1.3in]{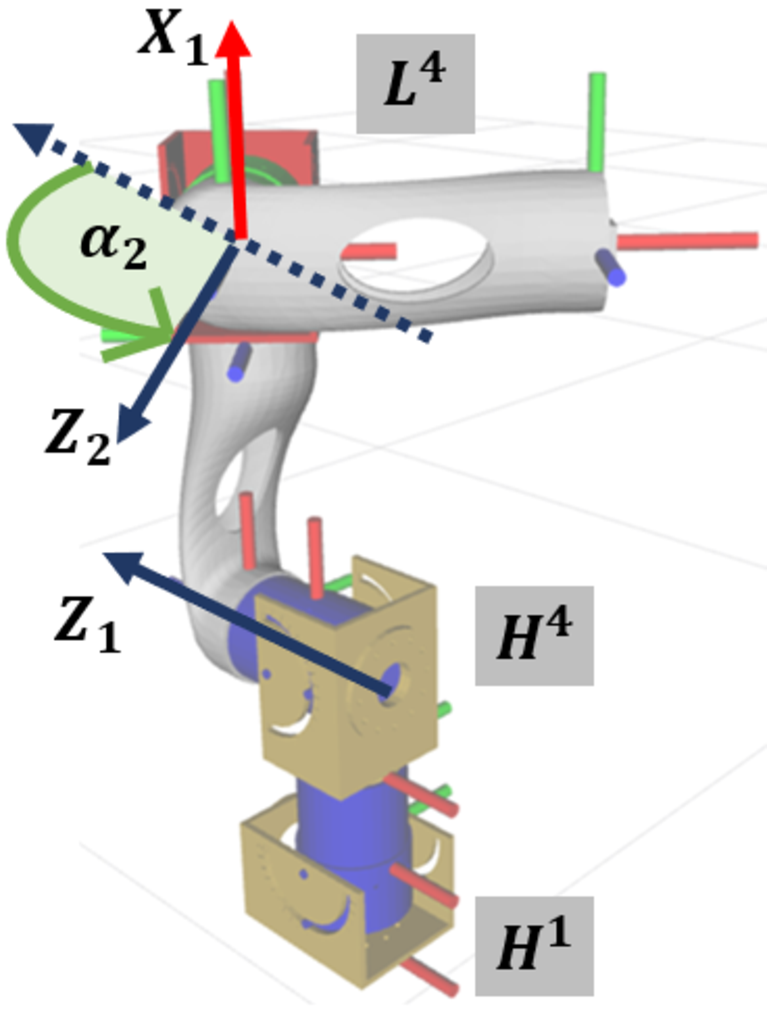}
	}
	~
	\subfigure[]{\includegraphics[width=1.65in]{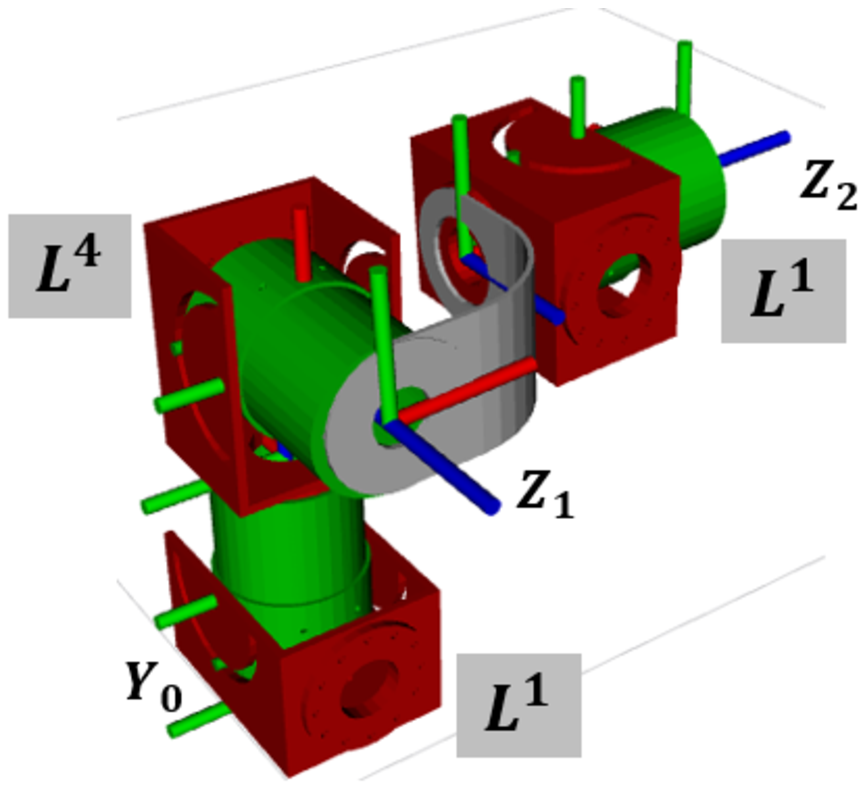}
	}
	~
	\subfigure[]{\includegraphics[width=1.4in]{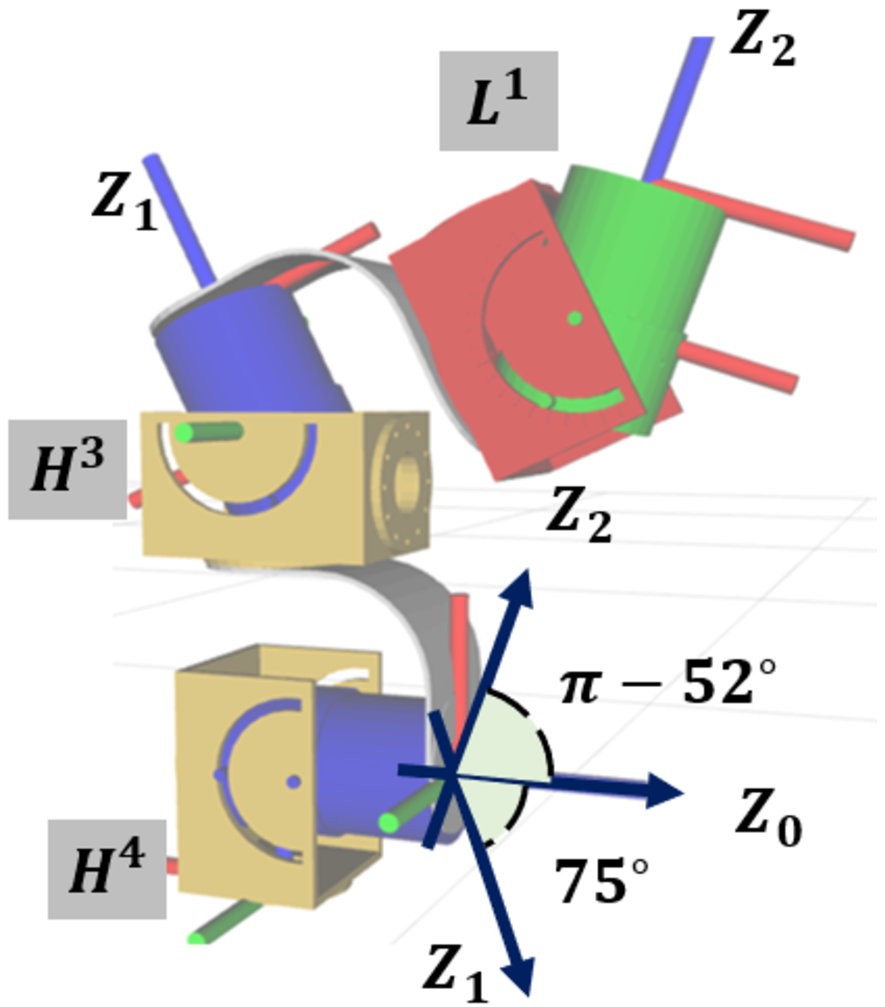}
	}
	\caption{(a)  Planar $H^3-L^4$, (b) Spatial $H^1-H^4-L^4$, (c) Spatial with twist (between $Z_1~\& ~Z_2$) $H^1-H^4-L^4$, (d) Wrist  $L^1-L^4-L^1$, (e) Raven-II $H^4-H^3-L^1$.}
	\label{fig:Types}
\end{figure}

\paragraph{Configuration type I:}\label{sec:types}
This is considered as the most common ones to see in manipulators with twist angles as $0^\circ$ and $90^\circ$. All the planar configurations and spatial configurations with these twist angles fall under this category and are the most commonly used due to the available analytical solutions of inverse kinematics. 
As shown in Fig.~\ref{fig:Types} (a), a $2-DoF$ planar configuration can be developed using two variants as heavy and light modules as $H^3-L^4$. For larger number of DoF in planar serial chains, $H^4$ would be repeating in sequence for $n-DoF$. Similarly, spatial configurations with $0^\circ$ and $90^\circ$ twist angles are shown in Fig.~\ref{fig:Types} (b) and (d). A 3-DoF configuration can be made with $H-H-L$ modular sequence using $H^1-H^4-L^4$ for reachable manipulators. Similarly $L-L-L$ can be used as $L^1-L^4-L^1$ for a spherical wrist configuration. 

\begin{table}
\caption{DH parameters and modular unit sequence of the different modular configurations.}
\label{tab:table_DH}
\begin{center}
\begin{tabular}{c c c c c c}
& & & & &\\ 
\hline
\hline
Configuration type & \multicolumn{4}{c}{DH parameters}& Modular Configuration\\
& $a$ & $\alpha$ & $d$ & $\theta$&\\
\hline
\hline
\multirow{2}{*}{I (a)} & $a_1$ &0 &0 & $\theta_1$ &\multirow{2}{*}{$H^3-L^4$} \\
& $a_2$ & $\pi$ & 0 & $\theta_2$&\\
\hline
\multirow{3}{*}{I (b)} & 0 & $-\pi$/2 & $d_1$ & $\theta_1$&\multirow{3}{*}{$H^1-H^4-L^4$}\\
& $a_1$ & 0 & 0 & $\theta_2$&\\
& $a_2$ & 0 & 0 & $\theta_3$&\\
\hline
\multirow{3}{*}{II (c)} & 0 & $-\pi$/2 & $d_1$ & $\theta_1$ &\multirow{3}{*}{$H^1-H^4-L^4$}\\
& $a_1$ & $\alpha_2$ & 0 & $\theta_2$&\\
& $a_2$ & 0 & 0 & $\theta_3$&\\
\hline
\multirow{3}{*}{I (d)} & 0 &$-\pi$/2 &0 & $\theta_1$ &\multirow{3}{*}{$L^1-L^4-L^1$}\\
& 0 & $\pi$/2 & 0 & $\theta_2$&\\
& 0 & 0 & 0 & $\theta_3$&\\		
\hline
\multirow{3}{*}{III (e)} & 0 &$\alpha_1$ &0 & $\theta_1$& \multirow{3}{*}{$H^4-H^3-L^1$}\\
& 0 & $\alpha_2$ & 0 & $\theta_2$&\\
& 0 & 0 & 0 & $\theta_3$&\\
\hline
\hline
\end{tabular}
\end{center}
\end{table}

\paragraph{Configuration type II:}\label{sec:type2}
Configuration type II possesses one or more links with unconventional twist angles, that is the values other then $0^\circ$ or $90^\circ$, as shown in Fig.~\ref{fig:Types} (c). Twist angles incorporated in this configuration are measured between the two skew (non-intersecting) joint axes of adjacent joints. The configuration type can be made same as in Type I using $H-H-L$ modular sequence with $H^1-H^4-L^4$,with a twist given in between $Z_1~\& ~Z_2$ about $X_2$.

\paragraph{Configuration type III:}\label{sec:type3}
Configuration type III also possesses unconventional twist angles with intersecting joint axes of the adjacent joints of any configuration. Fig.~\ref{fig:Types} (e) shows a similar configuration called RAVEN-II reported by Hannaford et al~\cite{hannaford2012raven}. It is customized for the medical surgery and has non-conventional values of twist angles. First pair of joint axes are intersecting each other at an angle $75^\circ$ and next pair is intersecting at an angle $\pi-52^\circ$. The proposed design of modules can be used for the development of such unusual customized configurations. Here, the modular sequence is $H-H-L$ and using $H^4-H^3-L^1$. \\
Assembly of the modular units validates the generation of desired configurations. The incorporation of twist angles and other robotic parameters are done within the modular units using three frames defined in section~\ref{sec:frames}. The next steps are to determine the kinematics and dynamics of the developed modular configuration, which are discussed in the following sections.

\section{Automatic and unified modeling of modular configurations}\label{sec:automaticmodel}
Modular architecture is designed to incorporate the desired parameters which indicates the geometrical adjustments between the frames. Given the proposed types of modules, a user can realize a desired configuration by planning a sequence of modular units to be used. This model can be used for further computational work only if the corresponding kinematic and dynamic modeling is in place, such as for generation, evaluation and optimization of modular compositions and for the motion planning and control of assembled composition. Therefore, two methods are presented for the model generation of the modular composition.
\subsection{Direct modular assemblies and generating respective models}\label{sec:direct}
An algorithm is developed, which takes sequence of modular-units, module-type (in Heavy (H) or Light (L)), corresponding angles with respect to the proposed frames as inputs, and provides the output file of the robot model in XML format, as shown in Fig.~\ref{fig:urdf}. 
This method is useful when the user has no prior information of the standard configuration parameters, such as, during the optimal synthesis of modular composition. The optimal synthesis is done either by enumerating all possible combinations and selecting the best out of it, or by evolving a modular sequence during the optimization itself while satisfying the given constraints. A database has been provided which contains the \textit{stl} file models of each of the components of the modules. The database contains the geometrical and the inertial data of the modular components. The modular components are designed using a professional design software called \textit{SolidWorks} and \textit{stl} files are exported from this. The developed virtual models are used not only to simulate the design in a given environment but also to run the developed configuration in realtime - using the controllers - to accomplish the required tasks.
\subsection{Conversion from standard designs to modular composition}\label{sec:Dh2modular}
The DH parameters are the most common standard convention which is used by the large robotic community. Therefore, an additional algorithm is proposed to convert the standard robotic parameters into the modular compositions. The DH parameters possesses the information of number of degrees-of-freedom (DoF) and relationship between the adjacent joint frames in terms of the position and the orientation. As per the discussions in section~\ref{sec:modularconfigurations}, modular configurations are formed using four modular units. Therefore, first step of the process is to interpret a given set of DH parameters and convert them into modular units sequence i.e. in terms of $H^k ~or~ L^k$ as shown in Table~\ref{tab:table_DH}. Along with that, unconventional twist parameters are to be incorporated within these modular units, if any. Followed by the rules of assembly given in section~\ref{sec:modularconfigurations}, the algorithm to convert DH parameters into modular unit sequence works upon the following points.
\begin{enumerate}
	\item If $a~\neq~0$ in the DH table, $H^3$ or $H^4$ are used.
	\item If $a=0~\cap~\alpha~\neq~0$ in the DH table, twist is given by rotating actuator casings about the pivotal axis in the adaptive twist unit of joint module.
	\item If $a~\neq~0~\cap~\alpha~\neq~0$ in the DH table, twist is given using $Ip_1$ or $Ip_2$ connections ports of the joint module.
	\item If $d~\neq~0$ in the DH table, $H^1$ or $H^2$ are used.
\end{enumerate}
Through this, the prescribed DH parameters are converted into the modular unit sequence. It is worthwhile to note that the desired configuration can also be realized by assembling the modules without considering the DH parameters. After reconfiguration, the transformation matrices defined in section~\ref{sec:frames} can be used to write the kinematic model as required. To compute the joint torques for any modular configuration, the inertial specifications of modular architecture are utilized to formulate a unified approach. The inertial parameters of the modular components are fetched from the modular sequence to represent them with respect to the joint frames, as described in section~\ref{sec:Dh2modular}. Euler-Lagrange equation can be used to formulate the equations of motion of the modular configurations~\cite{fu1987robotics}.

\subsection{Modular URDF: model representation of modular configuration}
\begin{figure}[t]
	\centering
	\includegraphics[width=4in]{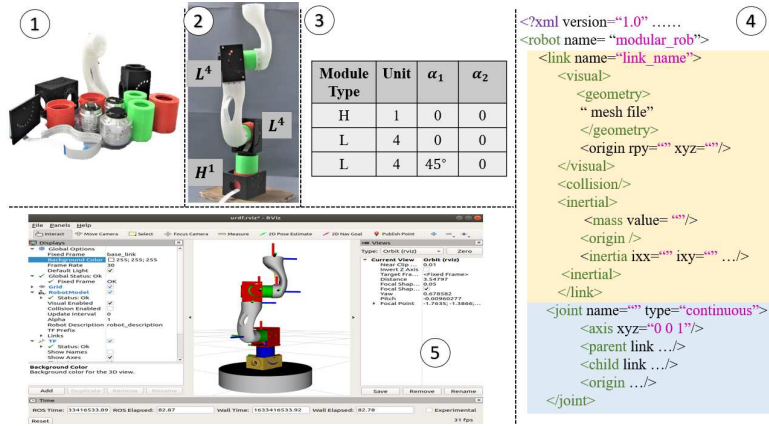}
	\caption{Flowchart for Unified Modeling: From, 1. modular library, 2. Assembled configuration, 3. Modular type-sequence, 4. Automatic URDF generation, 5. Visualization of configuration through Rviz.}
	\label{fig:urdf}
\end{figure}
For the automatic and on-the-fly modeling and control of a given modular configuration, the algorithm presented in section~\ref{sec:direct} and \ref{sec:Dh2modular} are implemented in python to generate the robot description files called as Unified Robot Description Format (URDF). URDF is widely used for the modeling and analysis of robotic systems in softwares like \textit{ROS, v-rep, OpenRave, Matlab} and so forth. URDF is an \textit{XML} (Extensible Markup Language) based file format to describe the various elements of a robot. The elements are specified in the form of links and joints tags, having sub-tags of visuals, origin, collision, inertia, etc., as shown in part 4 of the Fig.~\ref{fig:urdf}. Both the joints and the links are defined in terms of their rigid body geometrical and inertial parameters. Sensing elements, such as torque sensors, cameras, etc. can also be described in the same to include them in simulations and real time applications. URDF supports the tree like structures for robots, which provides an efficient method to determine the position and orientation of any joint/link based on the parent link or joint. Parent link is defined as the link which acts as a base for the next link. URDF is visualized in a debugger application called Rviz, through which generated models can be visualized and verified.\\
\subsection{Reconfigurable software architecture}\label{sec:software}
\begin{figure}[t]
	\centering
	\includegraphics[width=5in]{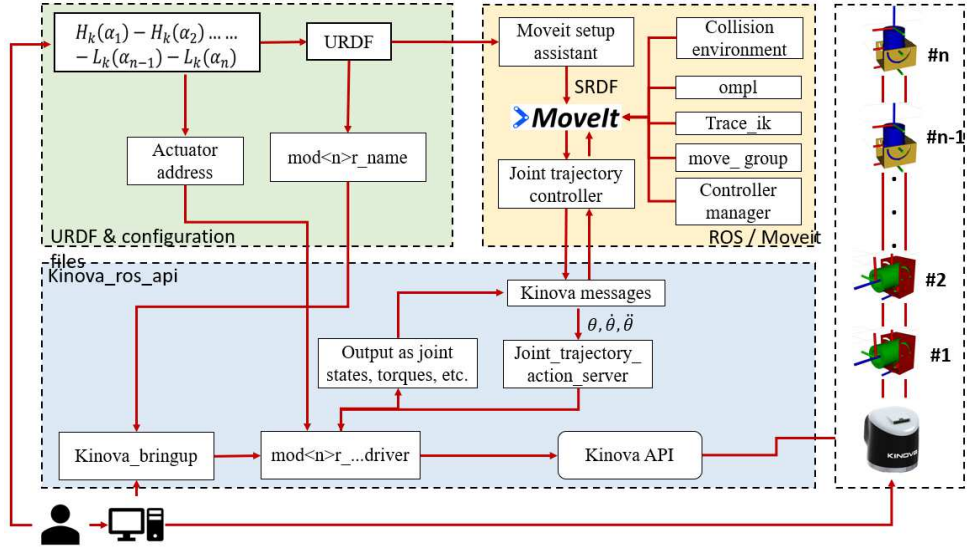}
	\caption{Software architecture and flowchart for motion planning and control of the modular compositions using proposed URDF generator, Moveit platform and Kinova API.}
	\label{fig:control}
\end{figure}

It is important to have reconfigurable software architectural system which can automatically adapt to the modular composition developed using the modular library. Every time the composition is disassembled and then assembled for a new one, with any DoF, the modular composition should be able to run without much effort, just like plug and play system. The proposed architecture is built upon the Robot Operating System (ROS), with the benefits of abstracting low level machine implementations such as joint control and communication, enabling the user to focus on higher level tasks, as shown in Fig.~\ref{fig:control}. The required Application Program Interfaces (API) are automatically configured right after the assembly of modules. 
Kinova actuators and controllers used in the development of modules are compatible with the ROS system. The software libraries, low-level API are provided by Kinova~\cite{kinova2019} for the actuators to connect them with ROS. The required changes in the configuration files are done automatically, as shown in Fig.~\ref{fig:control}. The name of the URDF file is to be set as $mod<n>\_name$, where $n$ defines the number of actuators to be used with actuator address stored in them. The actuator address actually specifies the joint number in the modular composition. Each variant of the actuator connects to each other in a daisy chain using a single flat flex cable which carry power and communication between modules. Starting from the base, this forms a chain of $n-$joint modules for $n-$DoF system. The actuator PIDs are used as-~$[2,~0,~0.05]$ which is working for all the unconventional modular configurations, as the system works upon the decentralized control scheme.

The URDF is converted to the Semantic Robot Description Format (SRDF) file using the moveit setup assistance package~\cite{moveit} in ROS, which adds the information about collision matrix, kinematic chains, joint composition, inverse kinematics plugin~\cite{tracik}, motion planning library~\cite{ompl} and type of controller~\cite{ros_control} to be used. For a given environment added in the interface, with assembled modular URDF, motion planning can be done between the given task locations. The planned motion commands are sent by the joint trajectory controller through kinova messages (definitions from API) to the Digital Signal Processor (DSP/base), which then sends the command to joint modules. User just have to launch certain files using a command line, such $kinova\_bringup$, $moveit\_launch$, and URDF generator as shown in Fig.~\ref{fig:control}. This is also shown through experimental demonstrations in the following sections. 

\section{Results}\label{sec:results}
The methodologies are implemented to generate $1-6$ DoF systems with all configuration types, using both the direct and conversion method to generate the modular composition in the virtual platform so as to replicate with real-scenario. The assembly rules, as briefed in section~\ref{sec:modularconfigurations}, are considered here to take care of the joint torque limits for the given payload of the configuration in study. Two case studies are presented in this paper considering different requirements of a designer.

\subsection{Modular configurations realization from given robotic parameters}
\begin{table}[h!]
	\caption{DH parameters for the case study of conversion from given parameters: Example A and B.}
	\label{table:CS1}
	\setlength{\tabcolsep}{8pt}
	\renewcommand{\arraystretch}{1.5}
	\begin{center}
		\begin{tabular}{c c c c c c}
			& & & & &\\ 
			\hline
			\hline
			\multirow{2}{*}{Example}& $a$  & $\alpha$ & $d$ & $\theta$ &\multirow{2}{*}{Modular Composition}\\
			& $(m)$& $(rad)$&$(m)$ & &\\
			\hline
			\hline
			\multirow{6}{*}{A} & 0 &$-\pi$/2 &0.148 & $\theta_1$ & \multirow{6}{*}{\includegraphics[width=1.2in]{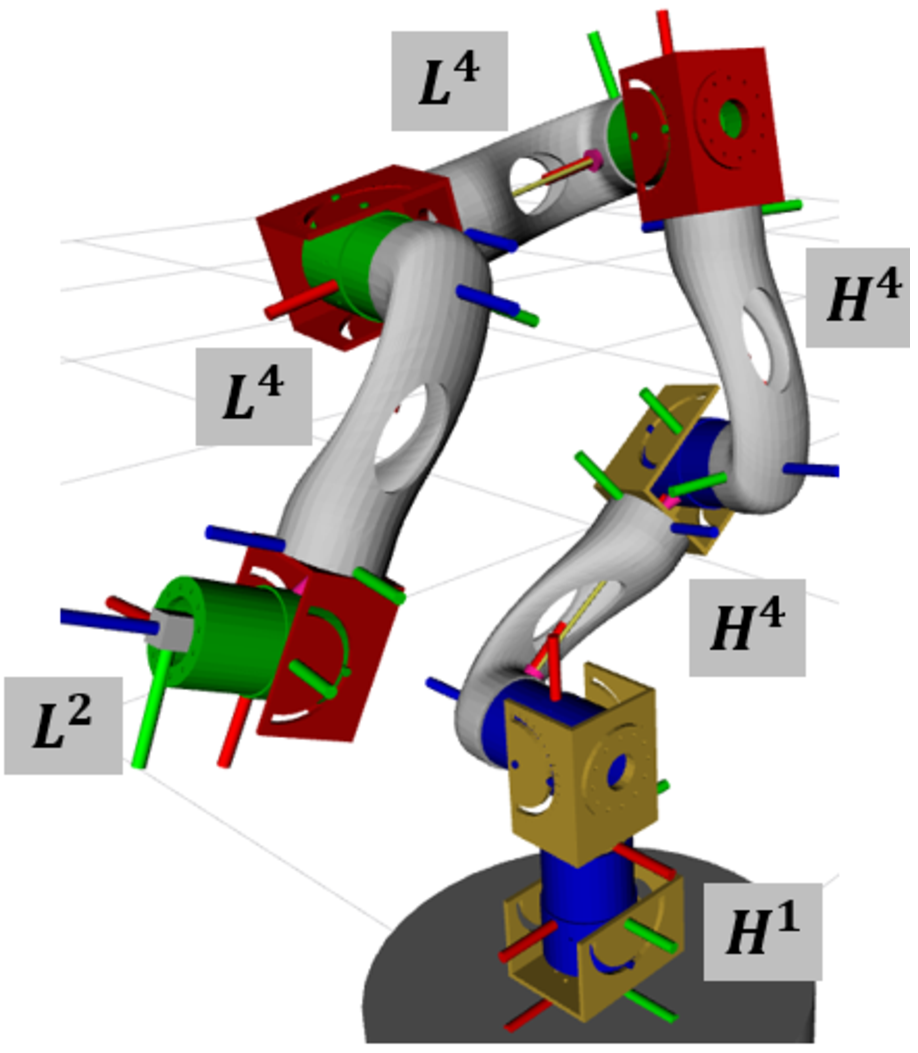}} \\
			& 0.3 & 0.26 & 0 & $\theta_2$&\\
			&0.3& 0& 0 &$\theta_3$&\\
			&0.3& 0.26& 0 &$\theta_4$&\\
			&0.3& $\pi$/2& 0 & $\theta_5$&\\
			&0&0 & 0.075 & $\theta_6$&\\
			\hline
			\hline
			\multirow{6}{*}{B} & 0&$-\pi/2$ &0.148 & $\theta_1$ & \multirow{6}{*}{\includegraphics[width=1.2in]{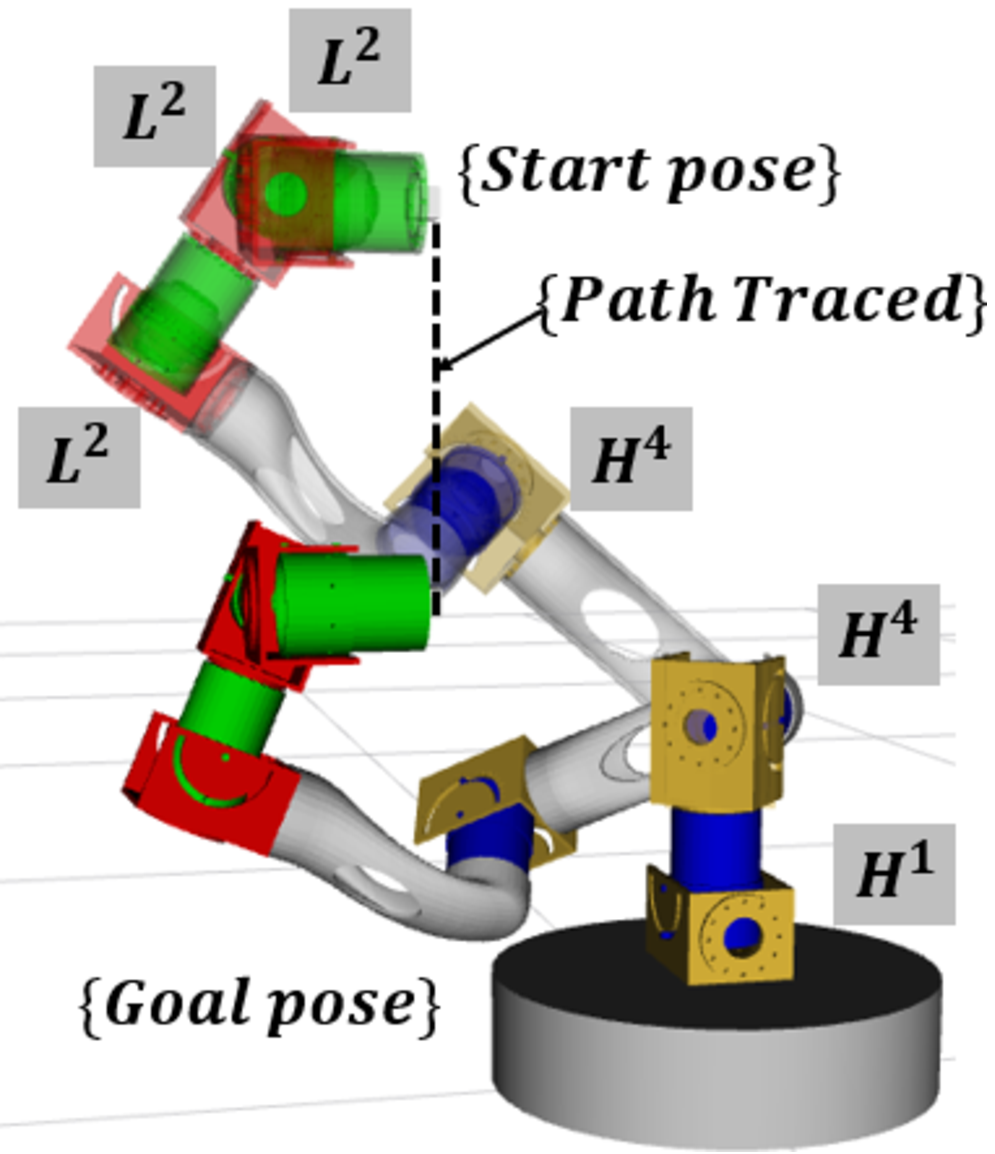}} \\
			& 0.3 & 2.36 & 0 & $\theta_2$&\\
			&0.3& 2.36& 0 &$\theta_3$&\\
			&0& $-\pi/2$& 0.148 &$\theta_4$&\\
			&0& $-\pi/2$& 0.148 & $\theta_5$&\\
			&0&0 & 0.075 & $\theta_6$&\\
			\hline
			\hline
		\end{tabular}
	\end{center}
\end{table}
For the task-based design, the configuration is generally synthesized using the optimization of the standard robotic parameters~\cite{singh2018modular}. Therefore, it is important to realize the standard parameters into the modular sequence for the given modular library. To showcase this, two exemplary cases are presented in which configuration can be realized using only the given robotic parameters. The examples are taken from the recent literature~\cite{singh2018modular, moulianitis2016task} who have shown the importance and the application of the unconventional configurations. The DH parameters of the candidate configurations are given in Table~\ref{table:CS1}. The modular sequence for the example A using the algorithm discussed comes out to be as $H^1-H^4-H^4-L^1-L^4-L^2$. This is an optimal configuration designed to work on predefined task space locations in a given cluttered environment~\cite{singh2018modular}. The modular sequence for the example B comes out to be as $H^1-H^4-H^4-L^2-L^2-L^2$. The configuration having skew-twists is optimized to follow a given rectilinear path~\cite{moulianitis2016task}. These results have shown the implementation of the conversion of robotic parameters into the modular sequence.

\subsection{Customized configuration generation and control in a given environment.}
\begin{figure}[h!]
	\centering

	\subfigure[]{\includegraphics[width=4in]{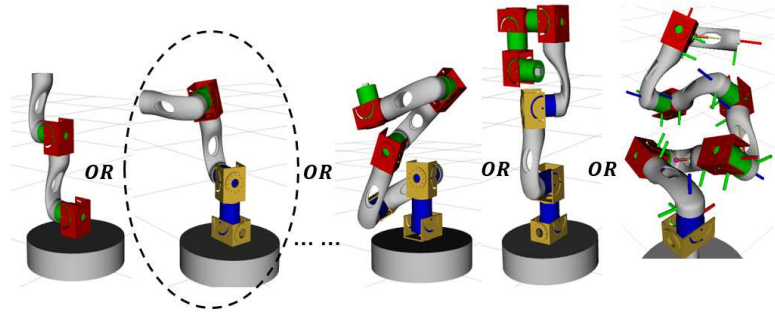}}

	\subfigure[]{\includegraphics[width=2.4in]{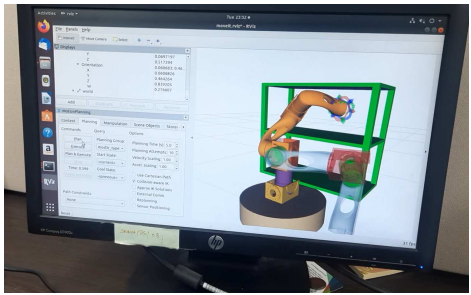}}
	~
	\subfigure[]{\includegraphics[width=2.15in]{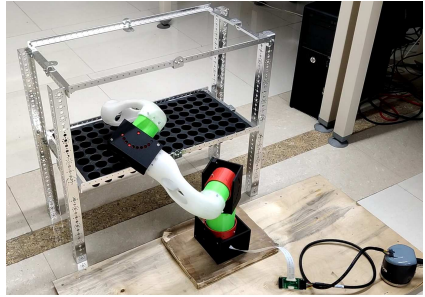}}
	\caption{(a) Selected configuration from optimal configuration results or from various possibilities after trails and validation, (b) Generated URDF and configuration files for motion planning and controls are integrated in ROS moveit: Real-scenario is replicated by the virtual platform.  (c) Deployment of custom modular configuration for the task in given environment:  3-DoF configuration reaching inside the vertical farm cell.}
	\label{fig:designer_process}
\end{figure}
For a given environment, as shown in Fig.~\ref{fig:designer_process} (c), it is required to develop a custom configuration to accomplish the given set of tasks. Since, a basic configuration is not fixed, it is challenging to select one to begin with. The designer can try out different configurations, analyze and validate them virtually using the developed platform. The initial step is to select the number and type of modules from a given modular library. Then, using the modular assembly rules from section~\ref{sec:modularconfigurations} and using modular units, see Fig.~\ref{fig:4types}, a modular assembly can be initialized which will automatically generate a URDF file for that particular configuration. This configuration can be visualized in ROS debugger called \textit{Rviz}, as shown in Fig.~\ref{fig:designer_process} (b). After trying out different configurations, which will take minimal time as compared to the physical verification, the selected configuration will be post-processed. 
Another technique is to generate/evolve an optimal configuration with respect to required performance measures~\cite{patel2015manipulator,icer2017evolutionary}. A 3-DoF configuration for a cell of the vertical farm set up, as shown in Fig.~\ref{fig:designer_process} (c) is developed using the proposed modular library, fabricated using a 3D printer with Poly-Lactic Acid (PLA) material, as shown in Fig.~\ref{fig:urdf}. This configuration is developed using $1-H$ module and $2-L$ joint modules. There is skew-twist angle of $45^\circ$ between the joint module 2 and joint module 3. The generated URDF is integrated in the motion planning package and control of the planned motion, as briefed in section~\ref{sec:software}. The generated URDF of this configuration is shown in Fig.~\ref{fig:designer_process} (a) with the integration of motion planning in Fig.~\ref{fig:designer_process} (b) and simultaneous motion in experimental setup.

Following the similar procedure, all other modular configurations can be developed, calibrated and tested for a given task. The proposed platform provides an on-the-fly validation, development and control of the modular and reconfigurable manipulators even with the unconventional parameters. The system as of now works based upon the decentralized control scheme and will be presented with the centralized controllers, model based adaptive controllers, in our future works.

\section{Conclusion}
Modules presented in this paper provides a customization approach through modularity and reconfigurability aspects which can also adapt the unconventional manipulator parameters. Configurations can be developed through modular assemblies or through conversion from design outcomes. In both the cases, kinematic and dynamic models are generated automatically. The framework provides an on-the-fly validation, development and control of the developed modular manipulators. Validation is done by automatic generation of the Unified Robot Description Format (URDF) of the unconventional modular configurations. Reconfigurable software architecture is presented for the motion planning and control of developed modular compositions. For the experimental validation, 3D printed modular library is used to assemble unconventional modular configurations and configured with ROS for motion planning and control in the real environmental setup and also validated through simulations.  
\bibliography{ms}

\end{document}